\documentclass[11pt]{article}

\usepackage[final]{acl}

\usepackage{times}
\usepackage{latexsym}

\usepackage[T1]{fontenc}

\usepackage[utf8]{inputenc}

\usepackage{microtype}

\usepackage{inconsolata}

\usepackage{graphicx}

\usepackage{titlesec}

\titleclass{\subsubsubsection}{straight}[\subsubsection]
\newcounter{subsubsubsection}[subsubsection]
\renewcommand\thesubsubsubsection{\thesubsubsection.\arabic{subsubsubsection}}
\titleformat{\subsubsubsection}
  {\normalfont\normalsize\bfseries}{\thesubsubsubsection}{1em}{}
\titlespacing*{\subsubsubsection}{0pt}{1.25ex plus 1ex minus .2ex}{1.25ex plus .2ex}
\usepackage{hyperref}

\makeatletter
\newcommand{\toclevel@subsubsubsection}{4}
\makeatother

\usepackage{todonotes}
\usepackage{subcaption}
\usepackage{wrapfig}
\usepackage{booktabs}
\usepackage{multirow}
\usepackage{mdframed}
\usepackage{minted}
\usepackage{amsmath}  
\usepackage{amssymb}  

\usepackage{float}
\usepackage{tabularx}
\usepackage{adjustbox}

\usepackage{forest}
\usetikzlibrary{arrows.meta, shadows}

\tikzset{
    base/.style={draw, align=center, font=\sffamily, rounded corners=2pt, drop shadow},
    level1/.style={base, fill=green!20, text width=2.5cm},
    level2/.style={base, fill=blue!20, text width=4.5cm},
    level3/.style={base, fill=pink!40, text width=4.5cm},
    example/.style={base, fill=yellow!30, text width=5cm, font=\sffamily\small},
    example2/.style={base, fill=yellow!30, text width=10cm, font=\sffamily\small},
    rootlabel/.style={font=\sffamily, rotate=90, anchor=center}
}

\definecolor{level0}{RGB}{200,200,200}
\definecolor{level1}{RGB}{181,220,113}  
\definecolor{level2}{RGB}{135,190,255}  
\definecolor{level3}{RGB}{252,176,183}  
\definecolor{citec }{RGB}{255,235,170}  

\title{Self-Improvement in Multimodal Large Language Models: A Survey}

\newcommand{\affmark}[1]{\textsuperscript{#1}}
\newcommand{\email}[1]{\texttt{\small #1}}

\newcommand{\twocol}[2]{%
  \parbox[t]{.48\linewidth}{\centering #1}%
  \hfill
  \parbox[t]{.48\linewidth}{\centering #2}%
}

\newcommand{\twocolw}[4]{%
  \parbox[t]{#1\linewidth}{\centering #3}%
  \hfill
  \parbox[t]{#2\linewidth}{\centering #4}%
}

\author{
\textbf{Shijian Deng}\affmark{1} \quad
\textbf{Kai Wang}\affmark{2} \quad
\textbf{Tianyu Yang}\affmark{3} \quad
\textbf{Harsh Singh}\affmark{4} \quad
\textbf{Yapeng Tian}\affmark{1} \\
[3pt]
\twocol{\affmark{1}The University of Texas at Dallas}
       {\affmark{2}University of Toronto} \\
\twocolw{.38}{.58}{\affmark{3}University of Notre Dame}
               {\affmark{4}\mbox{Mohamed bin Zayed University of Artificial Intelligence}} \\
[2pt]
\twocol{\email{\{shijian.deng,yapeng.tian\}@utdallas.edu}}
       {\email{kaikai.wang@mail.utoronto.ca}} \\
\twocol{\email{tyang4@nd.edu}}
       {\email{harsh.singh@mbzuai.ac.ae}}
}

\begin{document}
\maketitle
\begin{abstract}
Recent advancements in self-improvement for Large Language Models (LLMs) have efficiently enhanced model capabilities without significantly increasing costs, particularly in terms of human effort. 
While this area is still relatively young, its extension to the multimodal domain holds immense potential for leveraging diverse data sources and developing more general self-improving models. This survey is the first to provide a comprehensive overview of self-improvement in Multimodal LLMs (MLLMs). 
We provide a structured overview of the current literature and discuss methods from three perspectives: 1) data collection, 2) data organization, and 3) model optimization, to facilitate the further development of self-improvement in MLLMs. We also include commonly used evaluations and downstream applications. Finally, we conclude by outlining open challenges and future research directions.
\end{abstract}
    
\section{Introduction}
\label{sec:intro}

Self-improvement aims to enable models to collect and organize data required to build a better generation of themselves, which offers a path to overcome the costly scaling issues and potential performance ceilings of static training paradigms. In Multi-Modal Large Language Models (MLLMs), self-improvement seeks to use MLLMs themselves to obtain their own training data, resulting in improved MLLMs. Recent research~\cite{favero2024multi, deng2024enhancing, amirloo2024understanding} show that this approach can significantly reduce hallucinations and improve performance on general tasks with relatively low cost. Significant progress has been made in this direction. Some current studies~\cite{zhou2024aligning} partially leverage self-improvement by combining it with external tools or peer models, while others~\cite{yu2024rlaif} explore approaches that rely solely on a single model to handle all processes, toward full self-improvement. Although previous work~\cite{tao2024survey} has summarized the self-improvement in text-only LLMs and other surveys study the general scope of MLLMs~\cite{yin2024survey, zhang2024mm} or specific issues such as hallucinations~\cite{bai2024hallucination}, there is no comprehensive survey that focuses on these self-improvement methods for MLLMs. To fill this gap, we dedicate this paper to providing a comprehensive review of this area and identifying the challenges that need to be addressed.

\begin{figure}[t]
    \centering
    \includegraphics[width=0.45\textwidth]{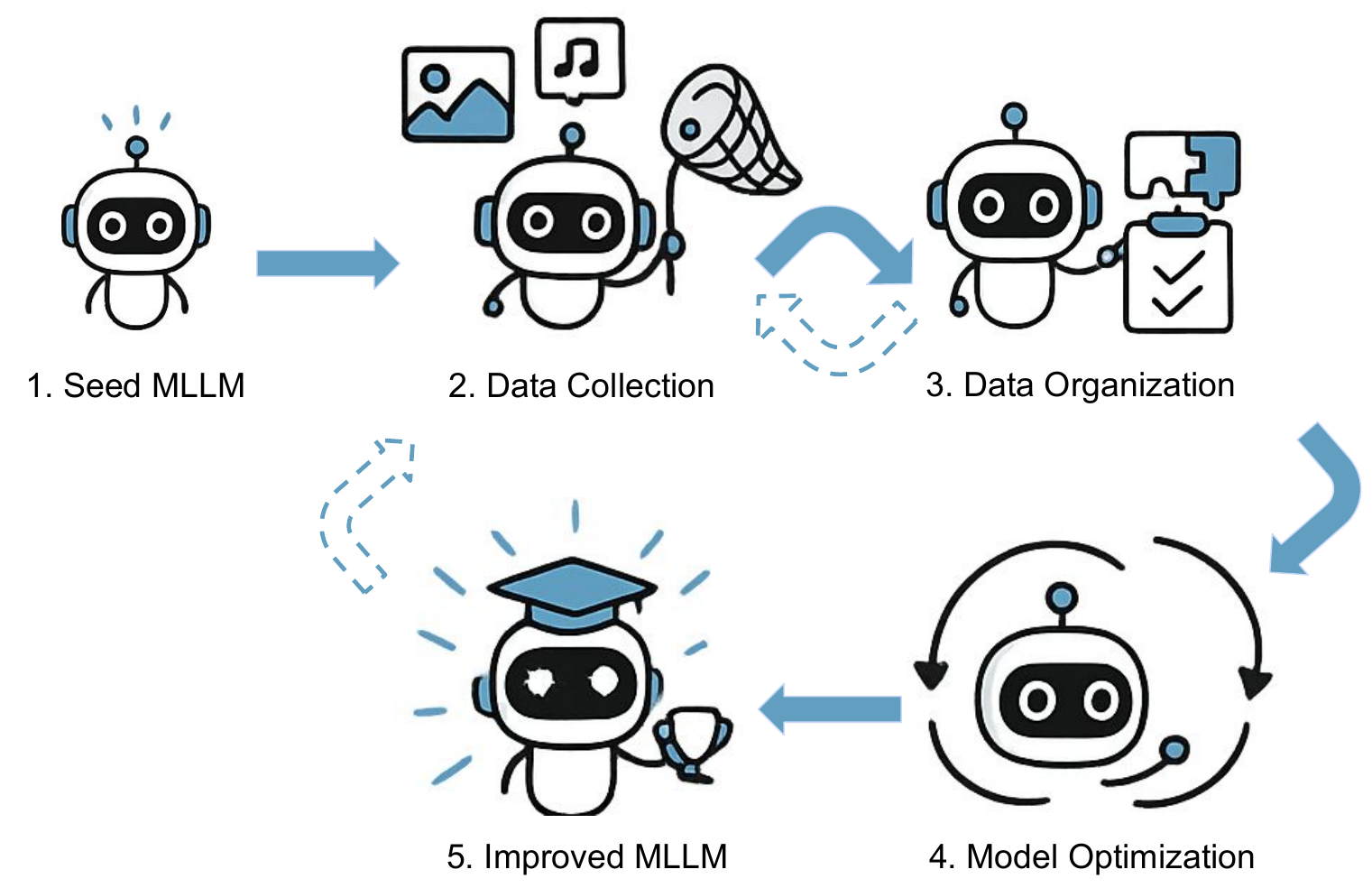}
    \caption{An illustration of self-improvement in Multimodal Large Language Models. The process involves selecting a seed MLLM to generate new data, organizing it into a dataset (which can optionally guide further data collection), and finally obtaining an improved model through training. This process can be iterated to achieve recursive self-improvement.}
    \label{fig:concept}
\end{figure}

Compared to self-improvement in LLMs~\cite{huang2022large, tao2024survey}, self-improvement in MLLMs faces unique challenges, such as the inclusion of multiple modalities. This can introduce modality alignment problems, which are known to cause issues like hallucination in MLLMs~\cite{li2023evaluating}. Additionally, MLLMs often cannot generate all the training data they need on their own, as most current models~\cite{liu2024visual, bai2023qwen} are unable to generate images directly.

Despite these challenges, there is growing interest in leveraging self-improvement in MLLMs to build models more effectively and efficiently. Promising results have already been achieved in this area. This paper aims to summarize previous works, compare methods, and provide clearer guidance for future research directions in this field.

\begin{figure}[t]
    \centering
    \includegraphics[width=0.4\textwidth]{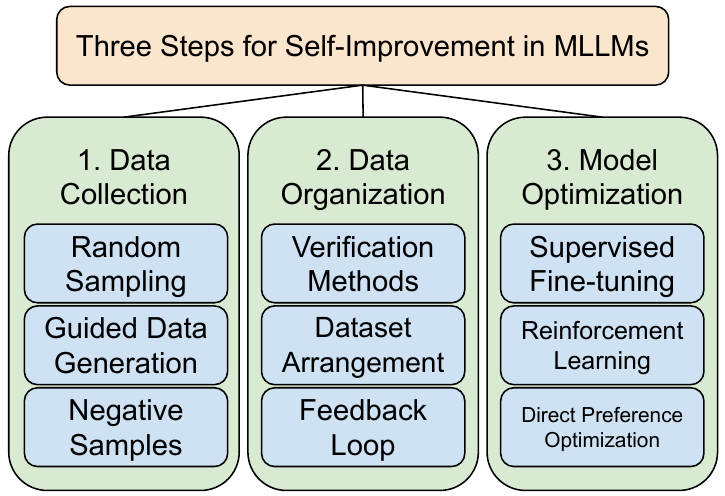}
    \caption{An overview of three steps for self-improvement in MLLMs. Each step can involve different methods based on requirements. For the full taxonomy please check Figure~\ref{fig:tax}.}
    \label{fig:overview}
\end{figure}

In this survey, we follow the structure outlined below: First, we provide an overview of the field. Next, we introduce the most commonly used seed models that serve as starting points for self-improvement. For the detailed methodology, we divide the discussion into three parts as shown in Figure~\ref{fig:overview}: data collection, data organization, and model optimization. We list current approaches and discuss their differences. We also collect evaluation methods commonly used to measure performance gains from self-improvement, compiling benchmark results for a comprehensive comparison. Additionally, we discuss downstream applications, to highlight the real-world impact of this paradigm. Finally, we identify the challenges in this field, which also represent potential future directions, and conclude the survey.

With this work, we aim to establish a clearer pathway for developing the next generation of MLLMs with better self-improvement mechanisms, moving beyond random exploration with biases. We hope to attract more researchers to explore this promising direction.

\section{Overview}
\label{sec:overview}

In this section, we first formally define self-improvement in multimodal large language models (MLLMs) in the context of this paper, and then compare it to similar concepts that have been used in MLLMs research. Afterwards, we summarize representative works in this domain to provide a general overview of the existing methods.

\subsection{Definition}

There are many similar terms to Self-Improvement, such as Self-Evolution, Self-Training, Self-Consistency, Self-Correction, Self-Reflection, and Self-Refinement, which have also been mentioned in previous MLLM research. There is a trend where the boundaries between these concepts are becoming blurred, and they may become more interchangeable in the future, depending heavily on the context. However, we clearly distinguish two paradigms. In this paper, we define self-improvement shown in Figure~\ref{fig:concept} as updating the model from $m_0$ to $m_1$, as opposed to self-refinement, which involves updating responses in context from $r_0$ to $r_1$. Formally, we express these concepts as follows:

\textbf{Self-Improvement (Model Update through Training):} \( m_1 = I(m_0, D) \), where \( I(\cdot) \) denotes the self-improvement operator that upgrades the entire model by training on self-curated multimodal dataset \( D \).

\textbf{Self-Refinement (Response Update in Context):} \( r_1 = R(r_0, c) \), where \( R(\cdot) \) represents the self-refinement operator that refines the initial response \( r_0 \) based on the context \( c \), which can be seen as a type of test-time scaling (or inference-time self-improvement~\cite{dong2024survey}). It is worth noting that some refined responses may have the potential to be incorporated into training data and thus contribute to further self-improvement.\footnote{Here, we do not consider storing newly acquired skills during inference in memory as an analogy for parameter tuning.}

A typical self-improvement process in MLLMs involves three modules: data collection, data organization, and model optimization as demonstrated in Figure~\ref{fig:overview}, which follows the structure of a general model-building process but focuses on automating the model development process using models rather than relying heavily on human intervention. While these commonly used modules are widely involved in the self-improvement of MLLMs, it is important to note that their life cycle does not necessarily conclude once an improved model is obtained. The iterative loop can persist, using the newly improved model as the seed for the next stage of self-improvement as demonstrated in Figure~\ref{fig:concept}. This life cycle can be highly dynamic, particularly in online settings, where data collection is directly influenced by the optimization design. This design may incorporate or encourage the model to explore more diverse or constrained data generation in subsequent rounds.

We conceptualize self-improvement in MLLMs as a spectrum of methods aiming to reduce human workload and maximize automation in improving model performance. Some methods target full autonomy, while others are limited to guided or assisted self-improvement, as long as they do not fully rely on human effort. Most papers in our survey do not leverage stronger external models. However, external models can be treated as tools that the seed model calls or uses. Under this formulation, we believe such approaches fall within the spectrum of self-improvement, albeit at the less independent end due to their reliance on external tools. To illustrate this, we add Table~\ref{tab:self_improvement_levels} comparing different levels of self-improvement in MLLMs, detailing what they automate and their limitations, allowing all discussed methods to fit organically within this spectrum.

\subsection{Related and Representative Works}

\begin{table*}[ht]
  \centering
  \tiny
  \begin{tabular}{@{} l p{7cm} p{5cm} @{}}
    \toprule
    \textbf{Level} & \textbf{Primary actor} & \textbf{Typical technique / example} \\
    \midrule
    L0 – No self-improvement
      & Humans do all data collection and curation
      & Flamingo~\cite{alayrac2022flamingo} \\
    L1 – Human-guided improvement
      & Model generates responses, while humans choose preferred data
      & RLHF-V~\cite{yu2024rlhf} \\
    L2 – Peer improvement
      & External models (e.g.\ GPT-4-V) supply data; minimal direct human effort
      & LLaVA~\cite{liu2024visual}  \\
    L3 – Hybrid self-improvement
      & Model collects its own data, but queries external augmentations or verifiers
      & Hybrid approaches, CSR~\cite{zhou2024aligning} \\
    L4 – Conditional self-improvement
      & Target model runs its own data loop except images are from existing datasets
      & RLAIF-V~\cite{yu2024rlaif} with self-reward \\
    L5 – High self-improvement
      & Model generates and curates both images and text without external data sources
      & SUDER~\cite{hong2025reinforcing}, UniRL~\cite{mao2025unirl} \\
    \bottomrule
  \end{tabular}
  \caption{Levels of multimodal self-improvement.}
  \label{tab:self_improvement_levels}
\end{table*}

Improvement without human supervision in MLLMs encompasses various strategies aimed at enhancing model performance through internal mechanisms. These approaches can be broadly categorized into Self-Refinement, Peer-Improvement, Self-Improvement for image LLMs, and extensions to Video LLMs and agents.

\subsubsection{Self-Refinement and Peer-Improvement}

Early methods like \textbf{Woodpecker}~\cite{yin2023woodpecker} and \textbf{VCD}~\cite{leng2024mitigating} focus on reducing hallucinations within generated content through training-free techniques. Due to the significant gap between proprietary models and early open-weight models, \textbf{LLaVA}~\cite{liu2024visual} and \textbf{HA-DPO}~\cite{zhao2023beyond} leverage GPT-4 to help build or refine multimodal capabilities, avoiding human supervision from scratch.

\subsubsection{Self-Improvement in Image Large Language Models}

Self-improvement strategies aim to enhance model abilities fundamentally by modifying model weights and reducing dependency on external models. Recent methods include on-the-fly enhancement of instruction-tuning data \textbf{VIGC}~\cite{wang2024vigc}, shifting from answering generation to self-questioning \textbf{SQ-LLaVA}~\cite{sun2025sq}, and synergy-driven cycles that interleave describing and locating objects \textbf{SC-Tune}~\cite{yue2024sc}. Others reduce hallucinations by converting training-free interventions into trainable ones \textbf{M3ID}~\cite{favero2024multi}, enabling interpretability in decision-making without extra annotations like \textbf{LLaVA-ASD}~\cite{deng2024hear}, leveraging data augmentation to construct preference pairs like \textbf{SeVa}~\cite{zhu2024self}, and applying step-wise self-rewarding \textbf{CSR}~\cite{zhou2024calibrated}. Some approaches rely on internal checks, such as visual metrics for preference tuning \textbf{SIMA}~\cite{wang2024enhancing} or using the model’s own encoder for fine-grained alignment \textbf{FiSAO}~\cite{cui2024fine}.

\subsubsection{Extensions to Video}

\textbf{i-SRT}~\cite{ahn2024srt} applies self-improvement in video large language models, addressing the issue of self-generated preferences that are linguistically plausible but not grounded in the visual content of the associated video.
\textbf{Video-STaR}~\cite{zohar2024video} adapts the STaR approach for the video domain, enabling the use of any labeled video dataset (such as Kinetics-700) for video instruction tuning.

\subsubsection{Multimodal Agents}

When augmenting MLLMs as agents and allowing them to act or even interact with each other, self-improvement enhances model performance across various tasks, including learning through self-play in image identification~\cite{konyushkova2025vision} or improving decision-making in games such as Blackjack and ALFWorld~\cite{zhai2025fine}.
\section{Seed Models}
\label{sec:seed}

A seed model does not need to be exceptionally strong, but it must clear a small set of capability floors that the self-improvement loop relies on. If these floors are missing, the model tends to generate low-quality data and the loop either stalls or collapses~\cite{hu2025multi}.

\textbf{Capability floors.}
Some skills are costly to "retrofit" purely from self-improvement and therefore should be present in the seed:

\begin{itemize}
    \item Basic visual grounding
    \item Robust text-in-the-wild handling
    \item Temporal aggregation for video
    \item Coherent reasoning traces (for reflection)
\end{itemize}

\textbf{Common choices.}
Several commonly used MLLMs have been adopted as seed models in self-improvement research:

\begin{itemize}
    \item \textbf{LLaVA}~\cite{liu2024visual}: As one of the earliest popular MLLMs, LLaVA has been widely used in MLLM self-improvement research due to its representativeness. The most commonly used versions are LLaVA-1.5 (7B and 13B). Some works, such as STIC and BDHS, utilize LLaVA-1.6.

    \item \textbf{Qwen-VL}~\cite{bai2023qwen}: Built on top of Qwen-LM, this model uses a three-stage training pipeline: Pretraining, Multi-task Pretraining, and Supervised Fine-tuning, to optimize its performance.

    \item \textbf{InstructBLIP}~\cite{daiinstructblip}: InstructBLIP introduces an instruction-aware Query Transformer that extracts informative features tailored to given instructions. It is trained on 13 datasets converted into an instruction-tuning format.

    \item \textbf{MiniGPT4}~\cite{zhu2023minigpt}: An early open-source effort to replicate the capabilities of GPT-4, MiniGPT4 aligns a frozen visual encoder with a frozen advanced LLM (Vicuna) using a single projection layer.

    \item \textbf{Video-LLaVA}~\cite{lin2023video}: It is commonly used as a seed model in video models. As its name implies, Video-LLaVA is similar to LLaVA but also fine-tuned on video datasets. It is designed for both image and video comprehension tasks.
\end{itemize}

Beyond these commonly used seed models, some works train their own seed models from scratch using a pretrained LLM to maintain more control over the entire process and address specific needs.

\section{Data Collection}
\label{sec:collection}

Effective data collection is crucial for enabling MLLMs to acquire and refine specific abilities. In conventional machine learning approaches, data collection typically relies on extensive human labor. This labor-intensive process, while effective, can be both time-consuming and costly, and is often limited by the availability and scalability of human resources.

\begin{table*}[ht]
\centering
\tiny
\begin{tabularx}{\linewidth}{lXX}
\toprule
\textbf{Method} & \textbf{Benefits} & \textbf{Drawbacks} \\ 
\midrule
Random Sampling~\cite{zhao2023beyond, yu2024rlaif} & Easy to use; works for any MLLM & May not be efficient; difficult to obtain samples with desired features \\[0.8ex]
Prompt-Guided Generation~\cite{wang2024vigc, fangvila} & Highly controllable; can generate almost any type of response & Requires significant human effort; difficult to scale \\[0.8ex]
Chain of Thought~\cite{zhai2025fine, zohar2024video} & Can generate long responses for reasoning tasks & Sometimes produces redundant or irrelevant reasoning steps \\[0.8ex]
Input Injection~\cite{zhou2024aligning, zhu2024self} & Can generate negative examples & Minor distortions may sometimes produce better examples than undistorted ones \\[0.8ex]
Sourcing from Multiple MLLMs~\cite{li2023silkie, xiong2024llava} & Ensures diversity in generated outputs & Requires additional effort to manage different models \\
\bottomrule
\end{tabularx}
\caption{Comparison of data collection methods.}
\label{tab:comparison_data_collection}
\end{table*}

In the context of self-improvement for MLLMs, a shift towards autonomous data collection is both desirable and increasingly feasible, thereby reducing the dependency on human intervention. This approach not only enhances efficiency but also enables continuous and scalable learning. We compare advantages and disadvantages of these methods in Table~\ref{tab:comparison_data_collection}.

\subsection{Random Sampling}
\label{sec:random_sampling}

The most straightforward method for autonomous data generation is random sampling~\cite{zhao2023beyond}, where the model generates data by sampling from its existing knowledge base without specific guidance. Although random sampling is simple to implement and can produce a diverse set of data, it has notable inefficiencies such as the generation of redundant or irrelevant data, which can waste computational resources and time.

\subsection{Guided Data Generation}
\label{sec:guided_data_generation}

To address the inefficiencies of random sampling, guided data generation techniques have been developed~\cite{cheng2024vision}. These methods employ predefined pipelines with carefully designed prompts to steer the model towards generating desired and high-quality responses. One prominent technique is Chain-of-Thought (CoT), which encourages the model to generate intermediate reasoning steps before producing a final answer. In order to further improve sample efficiency, some approaches adopt search-based methods such as beam search and Monte Carlo Tree Search (MCTS) and its variants~\cite{yao2024mulberry}.

\subsection{Negative Samples}
\label{sec:negative_samples}

Negative samples are essential for refining the model's ability to distinguish between correct and incorrect responses, thereby enhancing its overall accuracy and reliability. Various strategies have been explored to generate negative samples autonomously. \textbf{Poorly Designed Prompts}~\cite{deng2024enhancing}: Crafting ambiguous or misleading prompts can lead the model to generate suboptimal or incorrect responses. \textbf{Distorted Images}~\cite{zhou2024aligning}: Introducing visual distortions or noise into images challenges the model's visual comprehension capabilities. \textbf{Attention Masking}~\cite{amirloo2024understanding}: Manipulating the attention mechanism during the decoding process can result in responses that focus on irrelevant parts of the input. Additionally, the generation of negative samples can be finely controlled by altering the decoding path~\cite{deng2025efficient}, which produces responses that are less grounded in the visual context to the desired level, serving as effective negative examples for training.

Some methods utilize peer models for data generation (distillation), but implementing the same pipeline with the seed model itself may theoretically produce similar effects.

\section{Data Organization}
\label{sec:organization}

The data collected by MLLMs may not be directly suitable for feeding back into the models without further processing. To ensure the efficacy of self-improvement, a thorough verification and processing step is essential before leveraging the newly obtained data. The quality of this organization process is paramount, as it directly determines the robustness and reliability of the self-improvement mechanism in MLLMs.

\begin{table*}[ht]
\centering
\tiny
\begin{tabularx}{\linewidth}{lXX}
\toprule
\textbf{Method} & \textbf{Benefits} & \textbf{Drawbacks} \\ 
\midrule
Pre-assigned Labels~\cite{zhou2024aligning} & No extra effort required after data collection & Cannot handle complex cases \\[0.8ex]
Rule-Based Organization~\cite{yue2024sc} & Highly explainable & Not robust enough for novel samples \\[0.8ex]
Self-Evaluation~\cite{ahn2024tuning} & No additional reliance on external tools & Can suffer from model bias or hallucinations \\[0.8ex]
Judgment by External Verifiers~\cite{sun2024stllava} & Well-defined verifiers are highly robust & Some verifiers may incur significantly higher costs \\[0.8ex]
Feedback from Environment~\cite{zhai2025fine} & Robust and requires minimal additional effort & Many cases may be difficult to implement \\
\bottomrule
\end{tabularx}
\caption{Comparison of data verification methods.}
\label{tab:comparison_data_verification}
\end{table*}

\subsection{Verification Methods}
\label{sec:verification_methods}

The verification process can be a critical step during data organization and is usually implemented using either predefined rules or sophisticated models. Each method has its own advantages and limitations, which are discussed below. We also compare these methods in Table~\ref{tab:comparison_data_verification}.

\subsubsection{Rule-Based Verification}
\label{sec:rule_based_verification}

Rule-based organization involves applying predefined criteria to assess the quality and correctness of the generated data. This approach is straightforward and computationally efficient but may lack flexibility in handling diverse data scenarios.
\textbf{Majority Voting (Ensembling or Consensus):} The simplest approach compares multiple generated responses and selects the one with the highest frequency. While easy to implement, it may not always yield the best quality data, as the most frequent response might still contain inaccuracies or lack diversity.
\textbf{Ground Truth Alignment}~\cite{he2024self}: For datasets with established ground truths, the verification can involve cross-referencing the model’s output with the correct answers. For instance, in terms of the tasks requiring bounding boxes, an Intersection over Union (IoU) threshold can determine the acceptance of generated content~\cite{yue2024sc}. If the IoU score exceeds the predefined threshold, the content is deemed acceptable; otherwise, it can be discarded or flagged for further review. Alternatively, IoU can also be used as a reward function during RL training~\cite{liu2025visual}.

\subsubsection{Model-Based Verification}
\label{sec:model_based_verification}

Model-based organization leverages additional models to assess the quality of generated data. This method can provide more nuanced evaluations and modifications but may introduce additional computational overhead.
\textbf{Peer Model Evaluation}~\cite{hernandez2025improving}: Utilizing separate peer models to judge the quality of outputs can reduce bias and improve the reliability of the verification process. These models can provide independent evaluations, enhancing the overall robustness of data verification.
\textbf{Self-Critic Mechanism}~\cite{wang2024enhancing}: The MLLM itself can generate the rewards that evaluate the correctness and relevance of the data at various levels-token~\cite{cui2024fine}, sentence~\cite{zhou2024calibrated}, or output. This allows for more detailed assessments compared to rule-based methods.

\subsubsection{Verification from the Environment}
\label{sec:verification_from_the_environment}

MLLMs acting as agents that interact with their environment can also leverage environmental feedback for verification. The environment can be either the real world~\cite{guo2025improving, chen2025conrft} or simulated environments, such as games~\cite{zhai2025fine, konyushkova2025vision}.

\subsection{Dataset Arrangement}
\label{sec:dataset_arrangement}

The new data can be further post-processed by, for example, removing low-quality responses, fixing answers, normalizing formats, or storing the data by category. Depending on the goal, the final dataset can be created by:
\begin{itemize}
\item \textit{Filtering/Rejecting} unwanted data~\cite{liu2024diving};
\item \textit{Editing/Refining} outputs or rationales (e.g., through generator-corrector workflows and reflective self-correction), sometimes involving more complex procedures such as \textit{Topic-aware overwriting}, where errors are corrected within semantic clusters~\cite{wang2024vigc,zhang2024reflective,he2024self,he2024topic};
\item \textit{Archiving/Scheduling} hard examples for future rounds of curriculum learning as the model gets stronger~\cite{han2025self}.
\end{itemize}
The resulting dataset can be formatted for SFT or DPO, or it can be used to train a reward/judge model to guide future RL training or for other use cases such as evaluation~\cite{xiong2024llava}.

\subsection{Collection-Organization Loop}
\label{sec:co_loop}

The data collection-organization pipeline is not necessarily unidirectional. Data organized in early rounds can influence data collection in later rounds, creating a loop that iteratively enhances the dataset and, eventually, the model's performance.

\textbf{Iterative Evolution} (data-centric):
In each round, the current model generates data. An organization step then verifies, filters, or transforms this data into a curated set, which can be sent back to the original model to generate higher quality or more diverse data before being used for training. This process improves data quality and reduces noise over successive iterations \cite{luo2024mmevol,chen2025c2}.

\textbf{Recursive Improvement} (model-centric):
The loop also supports upgrading the model itself, so that the next round of data is produced and organized by a stronger model. This enables the co-evolution of data and model capability \cite{tan2024beyond,liu2024diving,deng2025openvlthinker,chen2025c2}.
\section{Model Optimization}
\label{sec:optimization}

After obtaining the organized dataset, the next step is to update the parameters of the seed model. Several training methods have been employed in self-improvement for MLLMs, including supervised fine-tuning, reinforcement learning, and direct preference optimization. As discussed in the paper DeepSeekMath~\cite{shao2024deepseekmath}, all these methods are actually connected. We compare advantages and disadvantages of these methods in Table~\ref{tab:comparison_model_optimization}.

\begin{table}[ht]
\centering
\tiny
\begin{tabularx}{\linewidth}{p{2.5cm}X X}
\toprule
\textbf{Method} & \textbf{Benefits} & \textbf{Drawbacks} \\ 
\midrule
SFT~\cite{wang2024vigc, luo2024mmevol, xiong2024llava}  
& Highly efficient when using existing high-quality datasets 
& Requires human effort or high-cost strong models \\[0.8ex]
PPO~\cite{yue2024sc, zhai2025fine}  
& A classic online RL method, easy to deploy 
& The reward model may be difficult to obtain \\[0.8ex]
GRPO~\cite{chen2025r1v} 
& More efficient than PPO since no value model is needed 
& Involves a trade-off between efficiency and the number of groups \\[0.8ex]
RFT~\cite{liu2024diving}  
& Can be used in an offline manner 
& All negative samples are discarded \\[0.8ex]
DPO~\cite{li2023silkie, ouali2025clip, luo2024probing}  
& Can leverage both positive and negative samples 
& May experience distribution shift issues after extensive training \\
\bottomrule
\end{tabularx}
\caption{Comparison of model optimization methods.}
\label{tab:comparison_model_optimization}
\end{table}

\subsection{Supervised Fine-tuning}
\label{sec:supervised_fine_tuning}

Instruction tuning, or supervised fine-tuning (SFT), has become a widely adopted post-training method to enable LLMs and MLLMs to follow instructions and solve a broader range of general tasks. In SFT, the model is trained to minimize the discrepancy between its predictions and the ground truth responses provided in the dataset.

Formally, given a dataset \( \mathcal{D} = \{(x_i, y_i)\}_{i=1}^N \), where \( x_i \) represents the input and \( y_i \) the corresponding target output, the objective is to minimize the cross-entropy loss:\vspace{-2mm}
\begin{equation}
\small
    \mathcal{L}_{\text{SFT}} = -\frac{1}{N} \sum_{i=1}^N \sum_{t=1}^{T_i} y_{i,t} \log p(y_{i,t} | x_i, y_{i,<t}; \theta)
\end{equation}\vspace{-2mm}
This loss function encourages the model to generate outputs that closely match the ground truth. In the context of self-improvement for MLLMs, it enables the new model to better align with the desired improvement goals in generated output.

\subsection{Reinforcement Learning}
\label{sec:reinforcement_learning}

Reinforcement learning (RL) methods have been used to improve MLLMs without human demonstration data, particularly for preference alignment and reasoning tasks. It aims to generate outputs that receive high rewards. The objective is then expressed as:

\begin{equation}
    \mathcal{L}_{\text{RL}}(\theta) = -\mathbb{E}_{(x,y)\sim D_{\pi_{\theta}}}r(x,y)
\end{equation}

Methods such as Proximal Policy Optimization (PPO) have been initially employed in RLHF for MLLMs~\cite{sun2023aligning}. More recently, GRPO~\cite{shao2024deepseekmath} has emerged as an efficient alternative to PPO for training MLLMs~\cite{chen2025r1v}, as it does not require a value model. They can also be used in Reinforcement Learning from Verifiable Rewards (RLVR).

\subsection{Direct Preference Optimization}
\label{sec:direct_preference_optimization}

Direct Preference Optimization (DPO)~\cite{rafailov2024direct} is a reinforcement-learning-free offline alternative for preference learning, which has become the de facto standard in preference optimization for MLLMs. Unlike SFT and Rejection Fine-Tuning (RFT), which can only leverage positive data, it can also take advantage of negative data. It formulates the optimization problem as follows:

Given a pair of outputs \( (y^+, y^-) \) where \( y^+ \) is preferred over \( y^- \), the objective is to maximize the likelihood of preferred outputs while minimizing the likelihood of dispreferred outputs. The DPO loss can be expressed as:\vspace{-2mm}
\begin{equation}
\small
    \mathcal{L}_{\text{DPO}}(\theta) = -\frac{1}{N} \sum_{i=1}^N \left[ \log \sigma(s(y_i^+) - s(y_i^-)) \right]
\end{equation}
This objective encourages the model to assign higher scores to preferred outputs compared to dispreferred ones.

\subsection{Other Enhanced Variants}
\label{sec:other_enhanced_variants}

Some works adjust the classic method by adding additional components~\cite{xiao2024detecting}, such as penalty terms for specific designs. For example, incorporating regularization terms can help maintain model stability and prevent overfitting to the preference data.

\subsection{Alternative Ways of Using Negative Samples}
\label{sec:alternative_ways_of_using_negative_samples}

It is worth noting that preference learning is not the only way to utilize negative data samples. Combining negative samples with self-reflection and correction using a CoT approach can further enhance model performance~\cite{cheng2024vision}. This involves generating detailed reasoning steps that allow the model to identify and correct its own errors, thereby improving the quality of the outputs.

\subsection{Curriculum}
\label{sec:curriculum}

Multi-stage training with different optimization methods has become common practice in MLLM training. Some research shows that certain training stages may hurt the model~\cite{zhou2025r1}, while other studies find that certain performance gains can be more easily obtained by combining different stages of optimization~\cite{huang2025vision}. Research has shown that even one training stage, such as RL, can benefit from being further divided into different substages~\cite{deng2025boosting}.
\section{Dataset and Evaluation}
\label{sec:data_eval}

Although some datasets created with the help of MLLMs can be used to further improve them, no benchmark has been specifically designed for self-improvement in MLLMs. Typically, researchers use existing MLLM benchmarks and report performance gains compared to the seed model and other SOTA models. Some attempts, such as LLM-Evolve~\cite{you2024llm}, aim to build a new type of benchmark; however, this particular benchmark operates in a non-parametric setting.

\subsection{Dataset}
Some datasets have been developed to facilitate the self-improvement of MLLMs.

VLFeedback~\cite{li2024vlfeedback} is the first large-scale, AI-annotated vision-language feedback dataset, containing over 82K multimodal instructions and comprehensive, model-generated rationales. The DeepPerception Dataset~\cite{ma2025deepperception} aims to enhance the cognitive visual perception capabilities of MLLMs for knowledge-intensive visual grounding (KVG); it comprises high-quality, knowledge-aligned training samples generated through an automated data synthesis pipeline. The OmniAlign-V-DPO Dataset~\cite{zhao2025omnialign} leverages answers from the OmniAlign-V SFT dataset as positive examples. To create the preference pairs necessary for DPO, negative samples are generated using another MLLM, LLaVANext baseline, through rejection sampling. The VisionPrefer Dataset~\cite{wu2024multimodal} is another high-quality, fine-grained preference dataset created to train a reward model for aligning text-to-image generative models. It aggregates feedback from AI annotators, specifically utilizing GPT-4V's capabilities to evaluate generated images based on defined criteria. The LLaVA-Critic dataset~\cite{xiong2024llava}, comprising 113,000 evaluation instruction samples across 46,000 images, was generated using a GPT-assisted pipeline, with GPT-4o providing judgment scores and reasons for evaluating MLLM responses. The community has also contributed additional datasets, including those used in RLAIF-V~\cite{yu2024rlaif} and Open-R1-Multimodal~\cite{EvolvingLMMs-Lab2025openr1multimodal}. These datasets created with MLLMs have demonstrated their utility in improving various MLLMs. We summarize these datasets in Table~\ref{tab:comparison_datasets} to highlight their differences.

\begin{table*}[htbp]
\centering
\tiny
\begin{tabularx}{\textwidth}{@{}llXX@{}}
\toprule
\textbf{Type of Dataset} & \textbf{Examples} & \textbf{How the dataset is created with MLLMs} & \textbf{How the dataset can be used for MLLMs} \\
\midrule

Instruction tuning dataset & DeepPerception (Stage-1) & A strong peer is used to generate CoT reasoning data (with ground truth provided in the input). & To train MLLMs with SFT to initialize cognitive-perceptual synergy for the target domain. \\
\midrule

\multirow{15}{*}{\parbox{2.5cm}{Preference dataset}} & VLFeedback & Instructions from many previous datasets; responses generated by a pool of LVLMs; preferences assigned by GPT-4V. & To train MLLMs with DPO to improve helpfulness and harmlessness. \\
\cmidrule(l){2-4}
& OmniAlign-V & Images from previous datasets; questions generated by GPT-4o (given prompts); responses are generated and refined by MLLMs; negative samples are generated by the LLaVANext baseline and filtered by an LLM judge. & To use SFT or DPO to align MLLMs with human preferences without decreasing general abilities. \\
\cmidrule(l){2-4}
& LLaVA-Critic & Pointwise data uses GPT-4o to provide judgment scores (input includes instructions, responses from previous datasets, and a GPT-4o response as a reference); pairwise data uses previous preference datasets with further judge's justification augmented by GPT-4o. & To train an MLLM judge with SFT to assign scores or rankings based on the prompt's criteria and to justify its judgments. \\
\cmidrule(l){2-4}
& VisionPrefer & GPT-4V is used to generate three types of feedback for generated images. & To train a reward model to score generated images. \\
\midrule

RLVR dataset & DeepPerception (Stage-2) & Images and prompts are reused from an existing dataset (while response text is sampled on the fly during training). & For RL with an IoU-based reward and a format reward to further improve performance on Knowledge-Intensive Visual Grounding. \\
\bottomrule
\end{tabularx}
\caption{Comparison of datasets created with MLLMs.}
\label{tab:comparison_datasets}
\end{table*}

\subsection{Benchmarks}

Evaluating the self-improvement of MLLMs can leverage current popular MLLM benchmarks. These benchmarks can be broadly categorized as follows:

\subsubsection{General Knowledge}
Benchmarks in this category assess the model's ability to understand and reason across multiple disciplines using multimodal inputs. Notable benchmarks include \textbf{MMMU}~\cite{yue2024mmmu} and \textbf{MMStar}~\cite{chen2024we}, which focus on comprehensive multimodal understanding across various academic and professional domains.

\subsubsection{Reasoning}
These benchmarks evaluate higher-order cognitive abilities and commonsense reasoning within multimodal contexts. Examples such as \textbf{Mathvista}~\cite{lu2023mathvista} and \textbf{VCR}~\cite{zellers2019recognition} are designed to test mathematical reasoning and commonsense understanding through visual inputs.

\subsubsection{Hallucination}
Detecting and mitigating hallucinations in generated content is crucial for reliable MLLMs. Benchmarks like \textbf{CHAIR}~\cite{rohrbach2018object}, \textbf{POPE}~\cite{li2023evaluating}, and \textbf{AMBER}~\cite{wang2023llm} provide metrics and evaluation frameworks to assess the accuracy and relevance of model outputs against visual inputs.

\subsubsection{Medical}
Medical benchmarks focus on the model's capability to understand and reason with medical images and related queries. Datasets such as \textbf{VQA-RAD}~\cite{lau2018dataset}, \textbf{SLAKE}~\cite{liu2021slake}, and \textbf{PathVQA}~\cite{he2020pathvqa} are designed to evaluate the model's proficiency in medical image analysis and question-answering tasks.

\subsubsection{Video QA}
Assessing MLLMs' understanding of dynamic visual content is addressed by video-based benchmarks. Notable datasets include \textbf{MSVD-QA}~\cite{xu2017video}, \textbf{MSRVTT-QA}~\cite{xu2017video}, \textbf{TGIF-QA}~\cite{jang2017tgif}, and \textbf{ActivityNet-QA}~\cite{yu2019activitynet}, which provide question-answer pairs based on video clips to test temporal and contextual reasoning.

\subsubsection{Judging Abilities}
Evaluating the model's capability to act as a judge involves assessing various aspects. Benchmarks like \textbf{VL-RewardBench}~\cite{li2025vl}, \textbf{MJ-Bench}~\cite{chen2024mj} are designed to measure these attributes, ensuring that the model's evaluations are reliable and consistent. Meanwhile, \textbf{AutoBench-V}~\cite{bao2024autobench} attempts to enable the MLLM itself to propose and construct new benchmarks.

\subsection{Meta-Analysis Across Benchmarks}
\label{sec:meta}
Using the compiled results, we observed the following robust patterns:

\paragraph{Method-Task Match.}
Rule-/verification-based RL (e.g., with step-wise or outcome checks) drives the largest absolute gains on verifiable tasks (visual math, programmatic reasoning, constrained captioning), while preference/AI-feedback data most reliably lowers hallucination metrics (e.g., POPE/AMBER) and improves general helpfulness/faithfulness.

\paragraph{Seed Strength Matters.}
Relative improvement $\Delta_{\text{seed}}$ typically shrinks as seed models get stronger; however, strong seeds show more stable gains across benchmarks. For identical pipelines (e.g., STIC-style), better seeds consistently yield higher end performance.

\paragraph{Cross-Benchmark Inconsistency.}
Methods that boost compositional reasoning can regress on perception-heavy tasks (fine-grained recognition, OCR, attribute binding), and vice versa. Pairwise rank correlations between benchmarks are often modest; gains on one suite do not guarantee gains on others.

\paragraph{Persistent Bottlenecks.}
We observe recurring difficulty in (i) fine-grained spatial reasoning (counting under occlusion, relative positions), (ii) multi-image/multi-hop consistency, (iii) long-horizon video temporal grounding, (iv) diagram/chart/plan understanding, and (v) robustness under noisy OCR or layout-heavy documents. Hallucination recurs in open-world scenes unless visual evidence is tightly verified.

\paragraph{Judge/Reward Leakage.}
When the same or closely related judges curate and evaluate (e.g., GPT-4V-like feedback used both for data construction and testing), scores inflate. Separation of curation and evaluation signals is critical for credible claims.

\paragraph{Efficiency Analysis.}
We discuss the efficiency of self-improvement methods in MLLMs from a computational cost perspective, considering factors like memory use during each stage and the data generation scale. First, regarding data sampling: random sampling often has the highest cost since it normally has a high rejection rate. Prompt-guided generation helps address this issue by giving more guidance, thereby reducing the search space of possible responses. Using negative samples further enables the usage of all generated data; even samples considered low-scoring can be used as negative samples, thus avoiding waste. For verification methods, the rule-based method generally has the lowest cost, since checking whether generated content satisfies rules is typically straightforward. Model-based verification can handle very complex scenarios but has higher cost. Verifying the outcome in the real environment can have the highest cost due to simulation complexity but may yield the highest feedback quality, especially for the most difficult verifications.

\section{Conclusion}
\label{sec:conclusion}

In this paper, we presented a comprehensive and structured survey of self-improvement in multimodal large language models (MLLMs). We defined the concept of self-improvement as used in this survey and clarified its differences from other related concepts. We discussed and compared representative works in this domain, highlighting their similarities and differences from three perspectives: 1) data collection, 2) data organization, and 3) model optimization. Further, we summarized commonly used evaluations and applications. Finally, we identified current challenges and potential opportunities for future research. We hope this survey serves as a valuable guide for researchers interested in exploring and developing new self-improvement methods for MLLMs.

\section*{Limitations}
\label{sec:limit}

Due to space limitations, this paper primarily focuses on a macro-level description and analysis of self-improvement within the current scope of MLLMs. Given the rapid evolution of the field, some of the most recent developments and new directions may not be included. Since we focus on the MLLM domain, we did not review work that involves only LLMs or agents; however, some methods may potentially be adapted to MLLMs as well. Despite these limitations, we believe this work, as the first survey in the area of self-improvement in MLLMs, provides a valuable overview of current research.
\bibliography{custom}

\appendix

\newpage
\section{Full Taxonomy}
\label{sec:full}

\begin{figure*}[t!]
    \centering
    \resizebox{\textwidth}{!}{%
        \begin{forest}
          for tree={
            grow=east,
            reversed=true,
            parent anchor=east,
            child anchor=west,
            l sep=20pt,
            s sep=10pt,
            edge={draw=black!60, thick},
            edge path={
              \noexpand\path[\forestoption{edge}]
              (!u.parent anchor) -- +(10pt,0) |- (.child anchor)\forestoption{edge label};
            }
          }
        [
          ,
          before drawing tree={
            !r.label={[rootlabel]left:Self-Improvement in MLLMs}
          }
          [1. Data Collection\\(§\ref{sec:collection}), level1
            [Random Sampling\\(§\ref{sec:random_sampling}), level2
              [{VLM-RLAIF\cite{ahn2024tuning}, RLAIF-V\cite{yu2024rlaif}, i-SRT\cite{ahn2024srt}, AnyPrefer\cite{zhou2024anyprefer}, FiSAO\cite{cui2024fine}, TPO\cite{he2024topic}}, example2]
            ]
            [Guided Data Generation\\(§\ref{sec:guided_data_generation}), level2
              [{VIGC\cite{wang2024vigc}, SQ-LLaVA\cite{sun2024sq}, SC-Tune\cite{yue2024sc}, Video-STaR\cite{zohar2024video}, VILA$^2$\cite{fangvila}, SCL\cite{he2024self}, R$^3$V\cite{cheng2024vision}}, example2]
            ]
            [Negative Samples\\(§\ref{sec:negative_samples}), level2
              [{POVID\cite{zhou2024aligning}, M3ID\cite{favero2024multi}, SeVa\cite{zhu2024self}, STIC\cite{deng2024enhancing}, BDHS\cite{amirloo2024understanding}, ESI\cite{deng2025efficient}, SENA\cite{tan2024beyond}, Image-DPO\cite{luo2024probing}}, example2]
            ]
          ]
          [2. Data Organization\\(§\ref{sec:organization}), level1
            [Verification Methods\\(§\ref{sec:verification_methods}), level2
                [Rule-Based Verification\\(§\ref{sec:rule_based_verification}), level3
                    [{SC-Tune\cite{yue2024sc}, BDHS\cite{amirloo2024understanding}, Video-STaR\cite{zohar2024video}, SCL\cite{he2024self}}, example]
                ]
                [Model-Based Verification\\(§\ref{sec:model_based_verification}), level3
                    [{VLM-RLAIF\cite{ahn2024tuning}, SIMA\cite{wang2024enhancing}, RLAIF-V\cite{yu2024rlaif}, i-SRT\cite{ahn2024srt}, STLLaVA-Med\cite{sun2024stllava}, LLaVA-Critic\cite{xiong2024llava}, R$^3$V\cite{cheng2024vision}}, example]
                ]
                [Verification from the Environment\\(§\ref{sec:verification_from_the_environment}), level3
                    [{RL4VLM\cite{zhai2025fine},
                    Visual\,-\,ARFT\cite{liu2025visualagentic}, MMSearch\,-\,R1\cite{wu2025mmsearch}, VRAG\,-\,RL\cite{wang2025vrag}, SEAgent\cite{sun2025seagent}, LASER\cite{wang2025learning}, iRe\,-\,VLA\cite{guo2025improving}, ConRFT\cite{chen2025conrft}}, example]
                ]
            ]
            [Dataset Arrangement\\(§\ref{sec:dataset_arrangement}), level2
              [Filtering \& Rejecting, level3
                [{M-STaR~\cite{liu2024diving}}, example]
              ]
              [Editing \& Refining, level3
                [{VIGC\cite{wang2024vigc}, RIT\cite{zhang2024reflective}, SCL\cite{he2024self}, TPO\cite{he2024topic}}, example]
              ]
              [Archiving \& Scheduling, level3
                [{Co\,-\,Improvement\cite{han2025self}}, example]
              ]
              [Judge Yielding, level3
                [{LLaVA\,-\,Critic\cite{xiong2024llava}}, example]
              ]
            ]
            [Collection-Organization Loop\\(§\ref{sec:co_loop}), level2
              [{MMEvol\cite{luo2024mmevol}, Beyond Human Data\cite{tan2024beyond}, M\,-\,STaR\cite{liu2024diving}, OpenVLThinker\cite{deng2025openvlthinker}, UniRL\cite{mao2025unirl}, SUDER\cite{hong2025reinforcing}, C2\,-\,Evo\cite{chen2025c2}}, example2]
            ]
          ]
          [3. Model Optimization\\(§\ref{sec:optimization}), level1
            [Supervised Fine-tuning\\(§\ref{sec:supervised_fine_tuning}), level2
              [{VIGC\cite{wang2024vigc}, SQ-LLaVA\cite{sun2024sq}, Video-STaR\cite{zohar2024video}, VILA$^2$\cite{fangvila}, MMEvol\cite{luo2024mmevol}, LLaVA-Critic\cite{xiong2024llava}, PVIT\cite{pi2024personalized}, R$^3$V\cite{cheng2024vision}, M-STaR\cite{liu2024diving}, Mulberry-7B\cite{yao2024mulberry}, VLM Dialog Games\cite{konyushkova2025vision}}, example2]
            ]
            [Reinforcement Learning\\(§\ref{sec:reinforcement_learning}), level2
              [{SC-Tune\cite{yue2024sc}, RL4VLM\cite{zhai2025fine}, M-STaR\cite{liu2024diving}, iRe-VLA\cite{guo2025improving}, ConRFT\cite{chen2025conrft}, Visual-RFT\cite{liu2025visual}, VisualThinker-R1-Zero\cite{zhou2025r1}, R1-Omni\cite{zhao2025r1}, Vision-R1\cite{huang2025vision}, Curr-ReFT\cite{deng2025boosting}, R1-Onevision\cite{yang2025r1}, R1-VL\cite{zhang2025r1}, Vision-R1\cite{zhan2025vision}, Video-R1\cite{feng2025video}, Skywork R1V\cite{peng2025skywork}, Skywork R1V2\cite{wei2025skywork}}, example2]
            ]
            [Direct Preference Optimization\\(§\ref{sec:direct_preference_optimization}), level2
              [{POVID\cite{zhou2024aligning}, M3ID\cite{favero2024multi}, SeVa\cite{zhu2024self}, HSA-DPO\cite{xiao2024detecting}, CSR\cite{zhou2024calibrated}, SIMA\cite{wang2024enhancing}, RLAIF-V\cite{yu2024rlaif}, AMP\cite{zhang2024automated}, STIC\cite{deng2024enhancing}, i-SRT\cite{ahn2024srt}, BDHS\cite{amirloo2024understanding}, CLIP-DPO\cite{ouali2025clip}, AnyPrefer\cite{zhou2024anyprefer}, SCL\cite{he2024self}, FiSAO\cite{cui2024fine}, ESI\cite{deng2025efficient}, TPO\cite{he2024topic}, SENA\cite{tan2024beyond}, Image-DPO\cite{luo2024probing}, SHAPE\cite{chen2025shape}}, example2]
            ]
          ]
        ]
        \end{forest}
    }

    \caption{The taxonomy of three steps for self-improvement in MLLMs. Each step can involve different methods based on requirements.}
    \label{fig:tax}
\end{figure*}
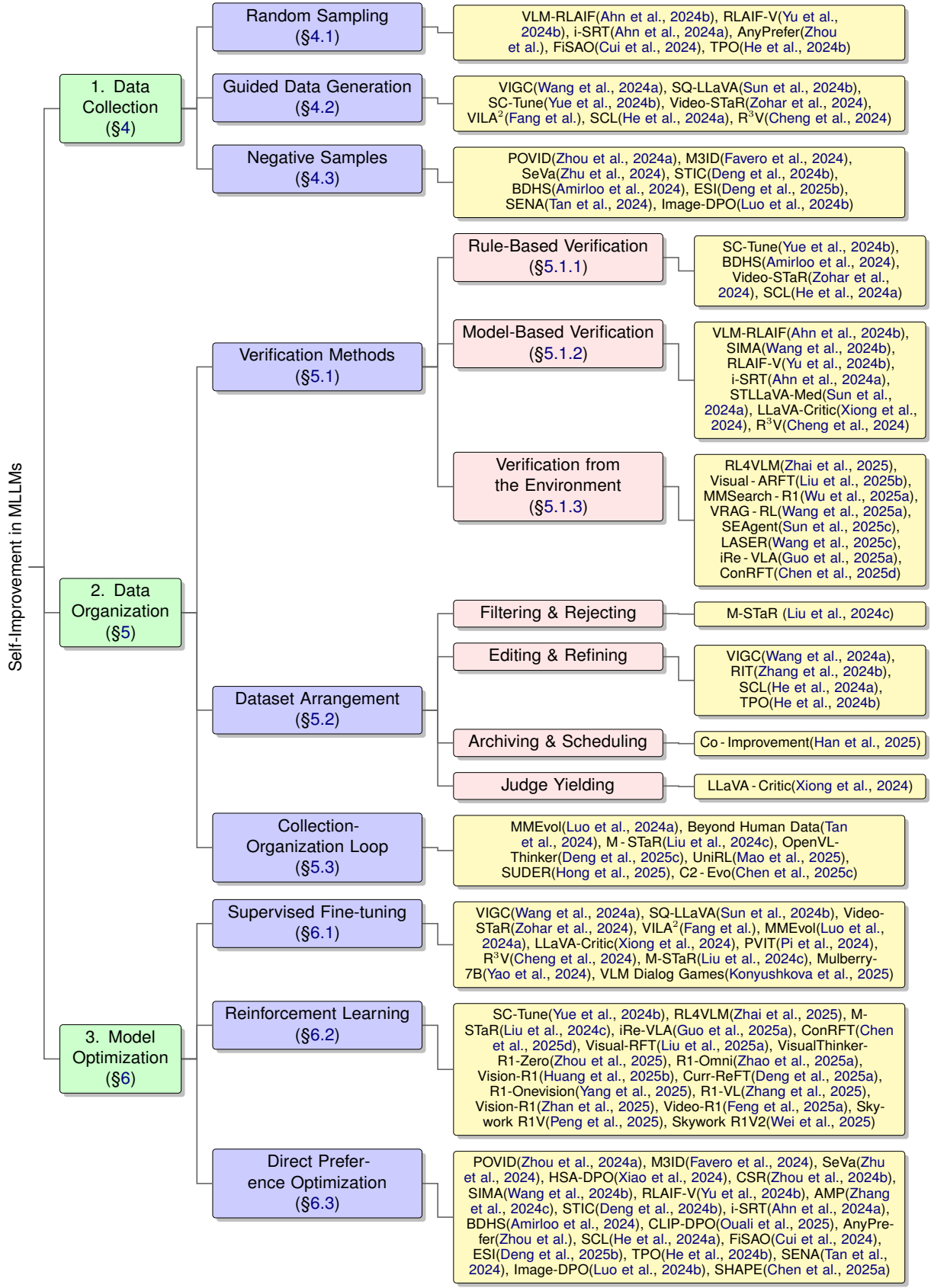

For space reasons, the main paper provides only an overview of our taxonomy of self-improvement in MLLMs. This appendix presents the complete hierarchy covering Data Collection (§\ref{sec:collection}), Data Organization (§\ref{sec:organization}), and Model Optimization (§\ref{sec:optimization}), and annotates each branch with representative works. See Figure~\ref{fig:tax} for the full diagram.

\section{Applications}
\label{sec:app}

Self-improvement can be particularly useful for applications that lack sufficient related instruction data. Models can autonomously generate the required data and conduct self-improvement to acquire new skills for downstream tasks.

\subsection{Math \& Science}
Tasks in fields like math and many other sciences require advanced reasoning sometimes including multimodal reasoning to address. However, the underlying reasoning data is not abundant, since humans seldom write down all the details of their reasoning steps, let alone reasoning that occurs via unconscious pathways. Self-improvement frameworks combined with peer-improvement have enabled MLLMs to autonomously generate and refine multimodal reasoning content, significantly reducing reliance on human-annotated data. For example, methods like MAVIS~\cite{zhang2024mavis} and COMET~\cite{liu2024comet} enhance mathematical reasoning by generating problems and visual explanations through structured prompts and alignment techniques. Similarly, frameworks like G-LLaVA~\cite{gao2023g} integrate geometry-specific tasks with generated datasets, achieving state-of-the-art performance on benchmarks like ScienceQA~\cite{lu2022learn}, SceMQA~\cite{liang-etal-2024-scemqa} and PHYSICS~\cite{feng2025physics}.

\subsection{Control}
Self-improvement in MLLMs can be applied to real-world applications such as control. Recent work~\cite{zhou2024anyprefer} proposes an automatic framework for preference data synthesis and employs an MLLM with an image segmentation model as a tool, judged by GPT-4o, to improve object segmentation and trajectory generation. The proposed method achieved a 15.50\% improvement in four visuo-motor control tasks.

\subsection{Healthcare}
Exciting advancements, such as STLLaVA-Med~\cite{sun2024sq}, have introduced the Self-Training Large Language and Vision Assistant for medical applications. This innovative approach focuses on training a policy model (an MLLM) to auto-generate medical visual instruction data, improving data efficiency through Direct Preference Optimization (DPO). Notably, a more robust and larger model (e.g., GPT-4o) serves as a biomedical expert, guiding the DPO fine-tuning process on the auto-generated data to effectively align the policy model with human preferences. This method achieves impressive zero-shot performance on three major medical VQA benchmarks: VQA-RAD, SLAKE, and PathVQA, while using only 9\% of the available medical data.
Additionally, LLaVA-ASD~\cite{deng2024hear} has explored using self-improvement approaches to enable MLLMs not only to assist in screening but also to provide explanations for their decision-making processes. This advancement offers a more explainable AI-assisted screening approach, enhancing transparency and user trust.

\subsection{Personalization}
With self-improvement approaches, users can easily personalize MLLMs~\cite{pi2024personalized, pham2024personalized} using automated pipelines to construct datasets and train models for their own use, requiring minimal additional effort.

\subsection{3D and Embodied Intelligence}
Recent advances in self-improvement for MLLMs also benefit areas such as 3D and embodied intelligence. A notable example is the MLLM-For3D framework~\cite{huang2025mllm}, which introduces a method for achieving 3D reasoning segmentation without the need for explicitly labeled 3D training data. This framework leverages pre-trained 2D MLLMs to generate multi-view pseudo segmentation masks along with corresponding text embeddings. These 2D masks are then projected into 3D space and aligned with the text embeddings, effectively transferring the 2D model's understanding to the 3D realm. Similarly, PiSA-Engine (Point-Self-Augmented-Engine)~\cite{guo2025pisa} has been introduced as a novel approach for generating instruction point-language datasets enriched with 3D spatial semantics. Streamlining Preference Alignment~\cite{jinspa}, a post-training stage designed for MLLMs equipped with 3D encoders, enhances the ability of MLLMs to understand and reason about 3D spatial relationships, which is fundamental for their effective application in 3D environments.

Self-improvement offers a powerful paradigm for enabling MLLM agents to improve their performance in embodied tasks through interaction with their environment. An example is SELU (Self-Learning in Unknown Environments)~\cite{li2024selu}, which allows MLLMs to improve their capabilities in embodied tasks without relying on explicit external human or environmental feedback. SELU adopts an actor-critic framework consisting of two MLLM components: the critic MLLM is responsible for evaluating the outcomes of the actor's actions and for improving its understanding of the environment. Simultaneously, the actor MLLM is improved based on the self-feedback provided by the critic. MART (MLLM As ReTriever)~\cite{yue2024mllm} is another example that enhances the performance of embodied agents by utilizing interaction data to fine-tune an MLLM retriever based on preference learning.

\section{Challenges and Opportunities}
\label{sec:challenges}

Self-improvement in MLLMs presents unique challenges and opportunities compared to text-only LLMs. We expand on these below:

\subsection{Uniqueness of Multi-Modality}

Many tasks and objectives in MLLMs fundamentally differ from those in LLMs. While LLMs primarily focus on maximizing the likelihood of text sequences, MLLMs must handle objectives incorporating spatial and temporal understanding. For instance, tasks involving images \(I\) or videos \(V\) require objectives beyond sequential prediction:
\begin{itemize}
    \item \textbf{Spatial Understanding (e.g., Object Detection):} Requires predicting bounding boxes \(B = \{b_k\}\) and classes \(C = \{c_k\}\). The objective might take the form:
    \begin{equation*}
    \mathcal{L}_{\text{spatial}} = \sum_{k} \left( \mathcal{L}_{\text{cls}}(c_k | I; \theta) + \lambda \mathcal{L}_{\text{reg}}(b_k | I; \theta) \right)
    \end{equation*}
    where \(\mathcal{L}_{\text{cls}}\) is a classification loss and \(\mathcal{L}_{\text{reg}}\) is a bounding box regression loss.
    
    \item \textbf{Temporal Understanding (e.g., Video Action Recognition):} Requires understanding sequences of frames \(V = (f_1, \dots, f_m)\) to predict an action \(a\). The objective could be:
    \begin{equation*}
    \mathcal{L}_{\text{temporal}} = - \log P(a | V; \theta)
    \end{equation*}
\end{itemize}

Cross-modal alignment and distillation without high-quality data~\cite{liu2024visual} might introduce multimodal hallucination. While text-only LLMs can hallucinate facts, MLLMs can hallucinate content inconsistent with an input image or other modality.

\subsection{Better Seed Models and Emerging Modalities}

Current self-improvement in MLLMs primarily operates on a limited set of modalities, typically \(\mathcal{M}_{\text{current}} = \{\text{Text}, \text{Image}, \text{Video}\}\). The action space \(\mathcal{A}\) for self-correction or data generation is often confined to textual outputs. However, significant potential lies in emerging modalities like Audio (\(A\))~\cite{wang2025self}, 3D data (\(D\)), and Embodied Actions (\(\text{Act}\))~\cite{ghasemipour2025self}, extending the modality set to \(\mathcal{M}_{\text{emerging}} = \mathcal{M}_{\text{current}} \cup \{A, D, \text{Act}, \dots\}\).

Expanding to these domains, particularly embodied AI, drastically increases the complexity and dimensionality of the action space. Self-improvement must transition from generating primarily discrete textual actions \(a \in \mathcal{A}_{\text{text}}\) to generating sequences of potentially continuous or high-dimensional actions \(a_t \in \mathcal{A}_{\text{embodied}}\) required for interaction within an environment \(E\). The optimization objective shifts towards maximizing expected return in sequential decision-making tasks:
\begin{equation*}
\max_{\pi_\theta} \mathbb{E}_{\tau \sim \pi_\theta} \left[ \sum_{t=0}^{T} \gamma^t R(s_t, a_t) \right]
\end{equation*}
where \(\tau = (s_0, a_0, s_1, a_1, \dots)\) is a trajectory generated by policy \(\pi_\theta\) in environment \(E\), \(s_t\) is the state (often multimodal), \(a_t \in \mathcal{A}_{\text{embodied}}\), \(R\) is the reward function, and \(\gamma\) is the discount factor. Works like~\cite{zhai2025fine, guo2025improving, chen2025conrft} are beginning to explore self-improvement in these expanded action and modality spaces.

\subsection{Omni I/O}

A limitation in MLLM self-improvement is the restricted input/output pipeline. Current models \(M\) often follow mappings like \(M: (\mathcal{M}_{\text{in}}, T_{\text{prompt}}) \rightarrow T_{\text{out}}\), where \(\mathcal{M}_{\text{in}}\) might be \(I\) or \(V\). Generating the non-textual input data (e.g., images \(I\)) often requires external datasets or separate generative models~\cite{luo2024probing}. This also limits MLLMs capabilities of self-verification and correction without extra models while forced to do so may compounding hallucinations.

True "Omni I/O" capability implies a model \(M_{\text{omni}}\) that can handle arbitrary combinations of modalities as both input and output. Let \(\mathbb{M}\) be the set of all relevant modalities. The mapping becomes:
\begin{equation*}
M_{\text{omni}}: \{m_{i}\}_{i=1}^{N_{\text{in}}} \rightarrow \{m'_{j}\}_{j=1}^{N_{\text{out}}}
\end{equation*}
where each \(m_i \in \mathbb{M}\) and \(m'_j \in \mathbb{M}\). For self-improvement, this means the model should ideally be able to generate its own training data across modalities, such as \(M_{\text{omni}}: T \rightarrow I\), \(M_{\text{omni}}: I \rightarrow T\), \(M_{\text{omni}}: (I, A) \rightarrow (T, V)\), etc., potentially in an interleaved manner. Recent advances like native image generation in GPT-4o/Gemini and open-source efforts like Qwen2.5-Omni~\cite{xu2025qwen2} suggest potential towards this goal, where self-improvement could enhance generation and understanding across text, vision, and audio within a single loop. Some work~\cite{qu2024silmm, zhao2025r1} has begun to unify these areas.

\subsection{Biases and Robust Verification}

After obtaining initial generated data, further verification and organization of this raw data are necessary, as we formulated these as the next steps for conducting self-improvement after collecting data. However, even with these controls, there is still no guarantee that bias and incorrectness can be eliminated. This is a significant challenge and an unsolved problem in self-improvement, as the bias may accumulate and potentially stop further recursive improvement, which presents a good opportunity for future research. The feasibility of self-improvement is intrinsically linked to the ability to reliably \textit{verify} the quality or correctness of the model's outputs. This echoes the computational complexity concept related to P vs NP: generating optimal outputs might be hard, but verifying them should ideally be tractable. We can formalize this with a verification function \(V(x, y)\), where \(x\) is the input and \(y\) is the MLLM's output (which could be multimodal). \(V(x, y)\) returns a score or a binary judgment (correct/incorrect, high/low quality).

Self-improvement often relies on optimizing parameters \(\theta\) based on this verification:
\begin{equation*}
\max_{\theta} \mathbb{E}_{(x, y) \sim P(x, y | \theta)} [V(x, y)]
\end{equation*}
or using \(V\) implicitly as a reward signal \(R\) in reinforcement learning. The core principle is: Effective self-improvement is contingent upon the existence of an efficient and reliable verification mechanism \(V\). If the complexity of verification, \(\text{Complexity}(V)\), is low (e.g., polynomial time), then iterative improvement guided by \(V\) becomes practical. As the real world is inherently multimodal, MLLMs could potentially leverage environmental feedback or cross-modal consistency checks as powerful verification signals~\cite{ahn2024tuning}, potentially making \(V\) more robust compared to text-only domains.

\subsection{Generalization}

Current self-improvement pipelines often focus on specific tasks \(\tau\) (e.g., reducing hallucinations, improving reasoning on benchmarks) and may exhibit diminishing returns after a finite number \(k\) of iterations:
\begin{equation*}
\theta_{i+1} = \text{Improve}( \theta_i, \mathcal{D}_i, \tau ), \quad i = 0, \dots, k-1
\end{equation*}
where \(\mathcal{D}_i\) is the data used/generated at iteration \(i\). Performance \(P\) might plateau, i.e., \(P(\theta_k, \tau) \approx P(\theta_{k+1}, \tau)\).

A major future direction is developing a \textit{general} MLLM self-improvement framework capable of recursive enhancement across a universal set of tasks \(\mathcal{T}_{\text{univ}}\) without plateauing. The idealized goal is a process:
\begin{equation*}
M_{i+1} = \text{SelfImprove}(M_i, \mathcal{T}_{\text{univ}}, \text{WorldKnowledge})
\end{equation*}
such that the model's capabilities \(C(M_i)\) monotonically increase across \(\mathcal{T}_{\text{univ}}\) as \(i \rightarrow \infty\):
{\small
\begin{equation*}
\forall \tau \in \mathcal{T}_{\mathrm{univ}},\;
\lim_{i \to \infty} P(M_i,\tau)
= \operatorname{OptimalPerformance}(\tau)
\end{equation*}
}
This requires mechanisms that not only refine parameters but potentially adapt the model's architecture, learning algorithms, and knowledge representation recursively, moving beyond narrow, task-specific improvement loops towards universal, open-ended capability growth.

\subsection{Scalability}

Although we have collected many models and frameworks in this survey, we found that many of these methods are normally conducted on a very small scale. Therefore, the performance gain is not as significant as in many other model developments that simply scale things up. It would be more practical and impactful for the real world deployment if the approaches had satisfactory scalability which would address the data shortage problem and therefore allow the model development to be further scaled up.

\subsection{Higher Autonomy}

Although current self-improvement MLLM frameworks can reduce the human workload from a data generation and verification perspective, human effort is still required in many other areas, such as proposing ideas, developing codebases, conducting experiments, and demonstrating or evaluating the final outcome. To overcome this bottleneck and achieve fully autonomous self-improvement requires higher autonomy, such as R\&D automation \cite{lu2024ai, yamada2025ai}. Meanwhile, these R\&D skills could themselves be further boosted by the improved base MLLMs, for instance, through a better multimodal understanding of the environment. This mutually beneficial self-improvement paradigm can increase effectiveness by removing bottlenecks, eliminating blind spots, and raising the upper bound.
\section{Related Surveys}
\label{sec:surveys}

There are several surveys on multimodal large language models (MLLMs)~\cite{yin2024survey} and self-improvement/evolution in LLMs~\cite{tao2024survey, he2025breaking}. However, to the best of our knowledge, no existing survey specifically addresses self-improvement in MLLMs. To fill this gap, we have collected related papers and systematically constructed this survey.

More recent works adjacent to our scope include (i) surveys on reinforcement learning for MLLMs~\cite{sun2025reinforcement, wu2025reinforcement}, which is a specific domain of self-improvement, and (ii) surveys on self-evolving agents that focus on agents rather than MLLMs~\cite{gao2025survey,fang2025comprehensive}.

Other surveys focus on topics such as self-supervised learning~\cite{gui2024survey}, self-training~\cite{amini2024self}, synthetic data~\cite{bauer2024comprehensive}, or data augmentation~\cite{feng2021survey}, which are loosely connected at a high level.

Our survey is the first to focus specifically on self-improvement in MLLMs, collecting a broad range of methods for automating MLLM improvement with less human effort. Concretely, we structure the field into a three-stage pipeline: data collection, data organization, and model optimization to analyze different techniques used in each module. We also formulate unified levels of autonomy for self-improvement in MLLMs to guide future development toward more effective self-improvement methodologies.

\end{document}